%% file: main.tex
\documentclass[11pt]{article}

\usepackage{acl}

\usepackage{times}
\usepackage{latexsym}
\usepackage{amsmath}
\usepackage{float}
\usepackage{amssymb}
\usepackage{enumitem}
\usepackage{makecell}
\usepackage{algorithm}
\usepackage{algorithmic}
\usepackage{diagbox}
\usepackage{booktabs}
\usepackage{array} 
\usepackage{tabularx}
\usepackage{multirow}
\usepackage{tcolorbox}
\usepackage{listings}
\usepackage[T1]{fontenc}

\usepackage[utf8]{inputenc}

\usepackage{microtype}

\usepackage{inconsolata}

\usepackage{graphicx}
\usepackage{CJKutf8}
\title{Bridging the Terminology Gap: A Data-Centric Alignment Framework for Multilingual Translation in Cross-Border Corporation}

\author{
  \textbf{Jinsheng Wei\textsuperscript{1,2}},\quad
  \textbf{Song Cheng\textsuperscript{1,}\thanks{~~Corresponding author.}},\quad
  \textbf{Bo Xiao\textsuperscript{1}},\quad
  \textbf{Ke Zeng\textsuperscript{1}} \\
  \\
  \textsuperscript{1}Meituan LongCat Interaction Team \quad
  \textsuperscript{2}Peking University \\
}

\begin{document}
\maketitle
\begin{abstract}
As globalization accelerates the demand for seamless cross-border enterprise collaboration, multilingual translation remains bottlenecked by insufficient domain-specific terminology, rigid terminology translation, weak instruction-following capabilities, and high inference latency in long-instruction scenarios. To overcome these barriers, we propose Bridging the Terminology Gap (BTG), a data-centric framework. First, BTG mines cross-lingual term alignments from domain documents and leverages LLMs to synthesize terminology-rich parallel corpora. Second, a hybrid filtering strategy combining heuristics and XCOMET scoring ensures high data fidelity. Third, a two-stage training paradigm employs supervised fine-tuning (SFT) for targeted terminology injection, followed by reinforcement learning (RL) to refine output preferences, effectively balancing domain knowledge integration with model reliability. Extensive offline evaluations (e.g., XCOMET, GEMBA, Acc@T) and online human assessments demonstrate significant improvements in terminology accuracy and instruction-following capabilities, validating the effectiveness of our framework, which is now deployed on our cross-border collaboration platform.
\end{abstract}

\section{Introduction}

\begin{figure}[t]
  \centering
  \includegraphics[width=1.0\columnwidth]{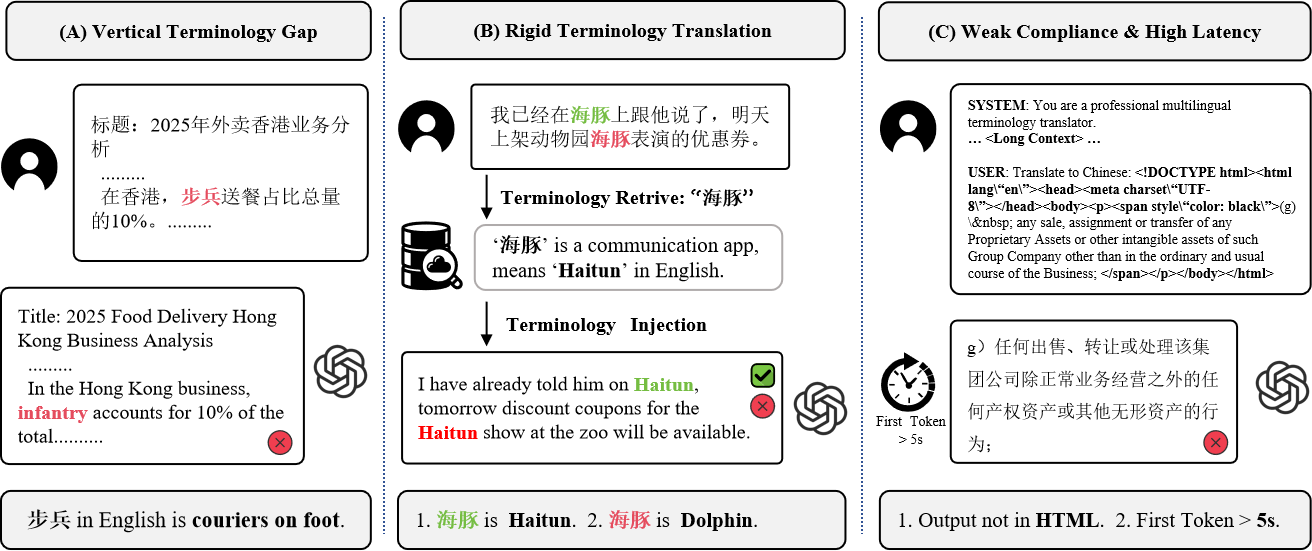}
  \caption{Illustration of critical challenges in cross-border collaboration translation.}
  \label{fig:introduction1}
  \vspace{-14pt}
\end{figure}

Globalized enterprise collaboration relies heavily on accurate multilingual translation. While Transformer-based Neural Machine Translation (NMT) \cite{fine} is the industrial standard for everyday scenarios \cite{10.5555/3295222.3295349}, LLMs have recently advanced specialized translation \cite{feng-etal-2025-mt-r1, wang-etal-2025-drt} through superior contextual awareness and reasoning capabilities. Despite these breakthroughs, handling complex vertical business scenarios in cross-border collaboration remains challenging:

\textbf{(1) Vertical Terminology Gap:} General LLMs prioritize broad fluency over domain specificity, often misinterpreting industry jargon. For instance, they erroneously translate the food delivery term ``\begin{CJK*}{UTF8}{gbsn}步兵\end{CJK*}'' as ``infantry'' rather than ``couriers on foot'' (Figure~\ref{fig:introduction1}.A).

\textbf{(2) Rigid Terminology Translation:} Prompt-based glossary injection imposes hard constraints that may override necessary contextual nuances. As illustrated in Figure~\ref{fig:introduction1}.B, enforcing a strict dictionary mapping forces the model to indiscriminately translate ``{\begin{CJK*}{UTF8}{gbsn}海豚\end{CJK*}}'' as the corporate app ``Haitun'' throughout the text, failing to recognize its second occurrence as the literal animal ``dolphin''.

\begin{figure*}[t]
  \centering
  \includegraphics[width=0.9\textwidth]{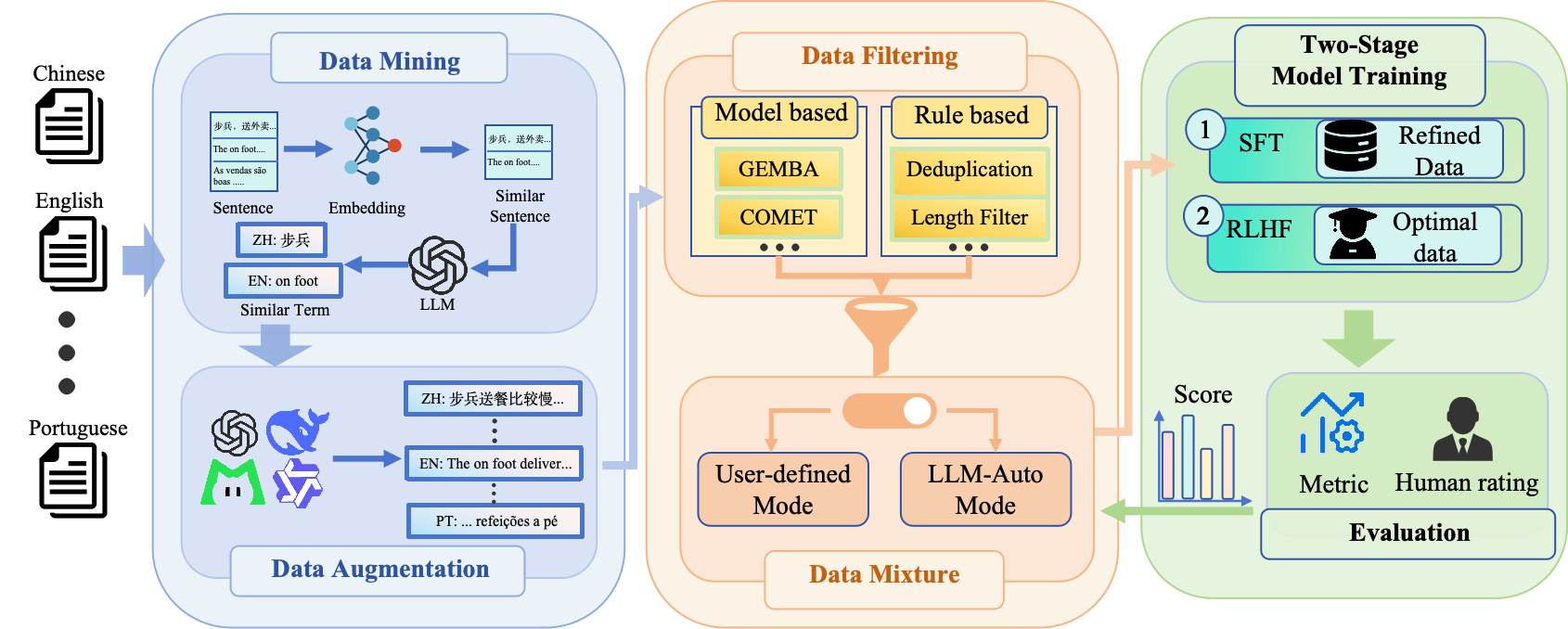}
\vspace{-8pt}
    \caption{Overall architecture of the BTG framework: data mining and augmentation(left), data filtering and mixing (middle), model training and evaluation(right).}
  \label{fig:framework}
\vspace{-20pt}
\end{figure*}
\textbf{(3) Weak Instruction-Following and High Latency:} Standard SFT often degrades instruction adherence, causing models to drop essential formats like HTML tags (Figure~\ref{fig:introduction1}.C). Conversely, relying on long-context prompts for knowledge injection introduces unacceptable Time-To-First-Token (TTFT) latency, severely disrupting real-time enterprise deployment.

To systematically address these challenges, we propose BTG (Bridging the Terminology Gap), a data-centric framework for vertical knowledge internalization. Our pipeline integrates automated terminology mining, LLM-driven augmentation, and hybrid metric-guided filtering to curate a high-fidelity corpus. This curated data drives a two-stage training paradigm, coupling SFT with RL, to effectively inject domain jargon while preserving generation fluency. Validated through a novel composite benchmark and successful online deployment, BTG significantly outperforms baselines in both terminology precision and overall quality.
To sum up, our main contributions are listed as follows:

\begin{itemize}

\item We identify the critical challenges of vertical terminology deficiency and the latency-reliability trade-off in cross-border collaboration, exposing current LLM limitations through real cases.
\item We propose BTG, a data-centric framework integrating automated terminology mining, metric-guided data filtering, a two-stage SFT+RL training paradigm to internalize domain knowledge and a comprehensive benchmark.
\item We construct a 5 million sample training corpus (blending general, terminology, and instruction data) and a comprehensive 30,000 example benchmark evaluated across 10 distinct metrics.
\item Extensive offline and online evaluations demonstrate BTG's superiority in term accuracy and fluency. Its large-scale enterprise deployment further validates its industrial viability.

\end{itemize}

\section{Related Work}
\input{related_work/related_work}

\section{Methodology}

\input{method/method}

\section{Experiments}

\input{experiment/experiment}

\section{Conclusion}
\label{sec:Conclusion}
In this paper, we have presented BTG, a data-centric framework designed to bridge vertical term gap in cross-border collaboration. By integrating semantic-aware term mining, metric-guided data filtering, and a two-stage training paradigm, our approach effectively internalizes domain knowledge into LLMs while preserving their general instruction capabilities. Extensive evaluations and online deployment show BTG achieves SOTA term accuracy and inference efficiency, providing a robust solution for global enterprise collaboration.

\section*{Limitations}

While BTG demonstrates robust terminology-aware translation capabilities, several limitations warrant future investigation. First, the two-stage training paradigm (SFT followed by RL) on a massive 5-million-sample corpus demands substantial computational resources, potentially limiting accessibility for resource-constrained deployments. Second, although our cross-lingual breakdown analysis confirms the framework's adaptability across the evaluated language pairs, its efficacy on extremely low-resource or morphologically complex languages remains unexplored. Finally, while we employ composite metrics to ensure rigorous evaluation, the partial reliance on LLM-based evaluators (e.g., the GM series) inevitably introduces subtle evaluator biases. Future work will explore lightweight alignment techniques and expand the terminology corpus to broader, highly specialized domains such as legal or medical translation.

\bibliography{custom}

\appendix

\section{Appendix}
\label{sec:appendix}

\subsection{Training Dataset Distribution}
\label{sec:training_dataset}
\input{appendix/training_dataset}

\subsection{LLM-Driven Automated Data Mixing}
\label{sec:llm_auto}
\input{appendix/llm_auto}

\subsection{Terminology Extraction Experiment}
\label{sec:term_extraction_appendix}
\input{appendix/term_extract_prompt}

\subsection{Cross-lingual Evaluation}
\label{sec:cross_lingual_Evaluation}
\input{appendix/cross_lingual_Evaluation}
\subsection{Inference Efficiency Experiment}
\label{sec:inference_efficiency}
\input{appendix/inference_efficiency}
\subsection{Training method ablation study}
\label{sec:training_method_albation_study}
\input{appendix/training_method_albation_study}

\subsection{Instruction Following Experiment}
\label{sec:instruction_appendix}
\input{appendix/instruct_follow}

\end{document}

%% file: related_work/related_work.tex

Traditional machine translation has evolved from statistical \cite{brown1990statistical} to neural approaches \cite{ijcai2020p496,ijcai2024p730}, yet domain adaptation remains a persistent challenge. Conventional solutions typically rely on in-domain fine-tuning \cite{do-etal-2023-unsupervised,du-etal-2020-adversarial}, constrained decoding with terminology banks \cite{ye-etal-2023-fid}, or dynamic adaptation mechanisms \cite{xu2024pluggable,nayak-etal-2023-instance}. However, these methods invariably require substantial parallel data, which is exceedingly scarce in corporate environments. 

While multilingual LLMs like mT5 \cite{mt5} and NLLB \cite{nllb} exhibit strong capabilities, cross-domain term extraction \cite{DiSTER2025} and adaptation remain problematic. Current adaptation strategies, such as prompt engineering \cite{prompt1,prompt2}, RAG \cite{rag1}, and fine-tuning \cite{10903789}, face severe enterprise deployment limitations, including prohibitive latency, rigid term insertion, and degraded instruction-following. Furthermore, although recent alignment methods utilizing RL \cite{tat,rl,POPostEdits2025} help resolve cross-lingual polysemy, they remain unoptimized for efficiency. This unresolved reliability-efficiency trade-off continues to hinder practical deployment in cross-border collaboration tools.

%% file: method/method.tex
This section details our proposed data-centric alignment framework for cross-border multilingual translation (Figure~\ref{fig:framework}). We sequentially introduce the corpus construction pipeline, which encompasses automated terminology augmentation (Section~\ref{sec:Automated Terminology Mining and Augmentation}) and metric-guided data mixing (Section~\ref{sec:Metric-Guided Data Filtering and Mixing}), before detailing the two-stage training paradigm (Section~\ref{sec:Two-Stage Knowledge Internalization}) and the final hybrid evaluation benchmark (Section~\ref{sec:Hybrid Human-Machine Evaluation}).

\subsection{Terminology Mining and Augmentation}
\label{sec:Automated Terminology Mining and Augmentation}
Vertical knowledge internalization hinges on the quality and coverage of domain data. To address the scarcity of parallel terminology data in cross-border corporation, we propose a two-phase framework consisting of semantic-aware data mining and contextual data augmentation.

\textbf{Semantic-Aware Terminology Mining.} To extract high-value parallel terminology pairs $\mathcal{T}$ from unstructured multilingual documents, we first segment the corpus and map sentences to dense vectors via a multilingual embedding model. By calculating cross-lingual cosine similarities and applying an empirical threshold ($\tau=0.7$), we isolate high-confidence aligned sentences. Subsequently, an LLM acts as a knowledge extractor to filter domain-specific jargon from these alignments, yielding precise translation tuples $(t_{src}, t_{tgt})$. See Appendix~\ref{sec:term_extraction_appendix} for prompt details and broader domain performance.

\textbf{Contextual Data Augmentation.}\label{sec:Contextual_Augmentation} Because relying solely on raw extracted terms risks overfitting to rigid structures, we implement an LLM-driven synthesis mechanism. For each term pair, we leverage LLMs to synthesize parallel sentences that seamlessly integrate the target jargon across diverse business contexts. This process yields an augmented dataset $\mathcal{D}_{aug}$ that expands the training scale while maintaining strict term consistency.

\subsection{Metric-Guided Data Filtering and Mixing}
\label{sec:Metric-Guided Data Filtering and Mixing}


\subsubsection{Data Filtering Strategy}
Our pipeline combines heuristic rules and model-based metrics to filter three data sources: general multilingual corpora, mined terminology (Sec.~\ref{sec:Automated Terminology Mining and Augmentation}), and instruction-following datasets.

\textbf{Rule-based Filtering.} A heuristic set $\mathcal{R}$ removes noise from candidate pairs $(x, y)$ by deduplicating, filtering text anomalies, enforcing strict language consistency, and applying a minimum length threshold $L_{min}$.

\textbf{Model-based Quality Scoring.} Surviving candidates are evaluated for semantic alignment and translation quality using LLM-based metrics (GEMBA-DA, MQM, ESA \cite{gemba,gemba-MQM}) and neural models (COMETKiwi, xCOMET-XL, xCOMET-XL-ref \cite{cometkiwi,xcomet}). A sample $(x, y)$ is retained only if its mean scores exceed pre-defined thresholds. The resulting high-quality dataset will be open-sourced.

\subsubsection{Adaptive Data Mixing Method}
\label{sec:Adaptive_Mixing}
The composition ratio of different data sources significantly impacts model performance. We developed an automated mixing tool that supports precise control over data types, sub-dataset ratios, total volume, and language priorities. This tool operates in two modes:

\textbf{User-Defined Mode.} This mode allows for flexible experimentation. Users can configure multiple sets of mixing hyperparameters $\boldsymbol{\lambda} = \{\lambda_{gen}, \lambda_{term}, \lambda_{inst}\}$, where $\lambda$ represents the ratio of each data type. The tool supports serial or parallel execution of training and evaluation pipelines based on these configurations.

\textbf{LLM-Driven Adaptive Mode.} To automate the search for optimal mixtures, this mode employs a feedback-driven loop, framing data composition as a hyperparameter tuning task. Here, an LLM acts as a policy optimizer that iteratively cycles through three steps: (1) training and evaluating the model using the current mixing ratio, (2) archiving the ratio and its resulting evaluation metrics into an experience bank, and (3) analyzing this historical trajectory alongside baselines to propose the next optimal ratio $\boldsymbol{\lambda}^{(t+1)}$. This iterative refinement continues until performance converges, ultimately yielding the best-performing data distribution (detailed in Appendix \ref{sec:llm_auto}).

\subsection{Two-Stage Knowledge Internalization}
\label{sec:Two-Stage Knowledge Internalization}

To achieve deep internalization of vertical knowledge while maintaining general translation and instruction following capabilities, we implement a two-stage training paradigm consisting of Supervised Fine-Tuning and Reinforcement Learning.

\textbf{Stage I: Supervised Fine-Tuning.} We optimize the model on the curated dataset $\mathcal{D}_{SFT}$ (Section~\ref{sec:Metric-Guided Data Filtering and Mixing}) via standard negative log-likelihood loss. This phase directly internalizes domain-specific terminology and vertical linguistic patterns into the model's parametric memory.

\textbf{Stage II: Preference Optimization via GRPO.} To enhance fluency and align with human preferences, we apply Group Relative Policy Optimization \cite{grpo}. Using a premium dataset $\mathcal{D}_{RL}$ (comprising top-scoring and expert-annotated samples), the model generates candidate groups $\{y_1, \dots, y_G\}$ for each input $x$, optimized under a custom composite reward function $R(x, y)$.

The total reward $R$ is composed of two components: a general translation quality score $R_{gen}$ and a terminology accuracy score $Acc@T$.
\begin{equation}
    R(x, y) = \alpha \cdot R_{gen}(x, y) + \beta \cdot {Acc@T}(x, y).
\end{equation}

(1) \textbf{General Translation Reward ($R_{gen}$):} To robustly evaluate translation quality across multiple dimensions, we aggregate scores from a set of $K$ metrics $\mathcal{M} = \{M_1, ..., M_K\}$ (selected from GEMBA, xCOMET, etc.). We adopt the \textit{Harmonic Mean} to penalize low scores in any single metric:
\begin{equation}
    R_{gen}(x, y) = \frac{K}{\sum_{k=1}^{K} \frac{1}{M_k(x, y)}}.
\end{equation}

(2) \textbf{Terminology Accuracy Reward ($Acc@T$):}  We define $\text{Acc@T}(x, y)$ to penalize term omission and reward precise injection. Given a term pair $(t_{src}, t_{tgt})$, the function assigns: (i) $+0.5$ when $t_{src}$ appears in $x$ and $t_{tgt}$ is correctly rendered in $y$, (ii) $-0.5$ when $t_{src}$ appears in $x$ but $t_{tgt}$ is missing from $y$, and (iii) $0.0$ otherwise.

Notably, for polysemous terms where rigid insertion would be semantically incorrect, the composite reward R naturally discourages over-application: such outputs incur a significant penalty from $R_{gen}$ via semantic quality metrics, even when Acc@T alone cannot distinguish them. By maximizing expected reward $\mathbb{E}[R(x, y)]$, model learns to generate translations that are not only terminologically precise but also linguistically fluent and preferred by comprehensive metrics.

\subsection{Hybrid Human-Machine Evaluation}
\label{sec:Hybrid Human-Machine Evaluation}

\subsubsection{Datasets and Benchmarks}
\label{sec:bechmark_datasets_eval}
We evaluate our framework on two dimensions of multilingual translation capability to ensure both breadth and depth:

\textbf{General Translation Domain:} To evaluate general translation capabilities, we construct a comprehensive benchmark of \textbf{20,829} samples, integrating Zh-En-Pt Parallel Chat (8,000), Corporate Encyclopedia (4,664), UN Documents (2,665), OPUS-100 (2,000), Literature and Film (2,000), Religion (1,200), and WMT24pp(300).

\textbf{Terminology Domain:} To assess cross-border cooperation translation, we curate \textbf{9,760} samples, comprising Internet Terminology (7,928) and Corporate Terminology (1,832).

\vspace{-8pt}

\subsubsection{Composite Automated Metrics}

To holistically evaluate both general fluency and domain precision, we define a base metric set $\mathcal{M}_{\text{base}}$ encompassing GEMBA (DA, MQM, ESA) and COMET (Kiwi, xCOMET-XL, xCOMET-XL-ref) variants, alongside Acc@T. Prioritizing the strengths of xCOMET-XL-ref, GEMBA-ESA, and Acc@T, we aggregate these normalized scores into two novel composite metrics to effectively balance overall translation quality with strict terminology adherence.

\textbf{(1) H-Ensemble (Holistic Harmonic Mean).} 
To penalize severe failure in any single dimension (e.g., fluent but wrong term), we propose H-Ensemble, which computes the harmonic mean of 7 key indicators with equal weight. Formally:
\begin{equation}
    \text{H-Ensemble} = \frac{7}{\sum_{m \in \mathcal{M}_{base}} \frac{1}{S_m} + \frac{1}{S_{\text{Acc@T}}}},
\end{equation}
where $S_m$ denotes the score of metric $m$. This metric serves as a "minimum barrel effect" indicator.

\textbf{(2) GA$_{\beta}$ (Focus-Aware Weighted Metric).}
In specific business scenarios, terminology correctness often outweighs general fluency. We propose GA$_{\beta}$, a weighted harmonic mean of GEMBA-ESA (representing semantic accuracy) and Acc@T. Similar to the $F_{\beta}$ score, the parameter $\beta$ controls the emphasis on terminology:
\begin{equation}
\text{GA}_{\beta} =\frac{ (1 {+} \beta^2) {\cdot} \text{GEMBA-ESA} {\times} \text{Acc@T}}{(\beta^2 {\times} \text{GEMBA-ESA}) {+} \text{Acc@T}}.
\end{equation}
A higher $\beta$ (e.g., $\beta=2$) places greater weight on Acc@T, making this metric ideal for strictly evaluating terminology compliance.

\subsubsection{Human Evaluation}
Beyond automated metrics, we incorporate human feedback through an online evaluation protocol. We deployed the trained model into  internal cross-border collaboration platform and conducted a blind side-by-side assessment on sampled business documents. Linguistic experts were tasked with evaluating the anonymized outputs based on multiple key dimensions: linguistic fluency, terminology accuracy, and adherence to cultural norms. The evaluation protocol consisted of two tasks: selecting the single \textit{Best Preference} and identifying all \textit{Acceptable Candidates} (multi-select).

%% file: experiment/experiment.tex
\input{experiment/main_exp_table1}

\subsection{Experimental Settings}
\textbf{Training and Evaluation Datasets.} Our training corpus comprises 5.54M rigorously cleaned samples spanning general translation, terminology, and instruction-following domains (see Appendix\ref{sec:training_dataset} for detailed distribution). For evaluation, we construct a 30k-sample \textit{Hybrid Multilingual Dataset} (Section \ref{sec:bechmark_datasets_eval}), mixing general and domain parallel sentences across 7 key cross-border languages (ZH, EN, RU, ES, PT, AR, FR).

\textbf{Evaluation Metrics.} We employ automated metrics across 3 dimensions: General Translation Quality, Vertical Terminology Accuracy, and proposed Composite Scores (detailed in Section \ref{sec:Hybrid Human-Machine Evaluation}).

\subsection{Main Results}
We benchmark BTG-trained Qwen3-8B (reward $\alpha=0.4, \beta=0.6$) against leading open and closed source models on a 30k-sample test set. As Table~\ref{tab:term_eval} details, our model yields state-of-the-art performance across CM series (CMKiwi: 82.38, xCM-XL: 88.57, xCM-XL-ref: 91.35), indicating superior fluency and expert alignment. In GM series, it trails top-performing GPT-4.1 by merely 4\%---a marginal gap plausibly reflecting inherent evaluator bias, given these metrics natively utilize GPT.

Crucially, BTG achieves the highest Acc@T (93.97) and BLEU~\cite{bleu}, successfully bridging the terminology gap while enhancing lexical ordering. Superior composite scores (H-Ensemble: 75.46, GA$_{\beta}$: 92.43) empirically validate an optimal equilibrium between general linguistic competence and strict terminology enforcement. Furthermore, a cross-lingual breakdown analysis of this main experiment (Appendix~\ref{sec:cross_lingual_Evaluation}) confirms our strong multilingual adaptability, consistently outperforming baselines in both semantic fluency (xCM) and terminology accuracy.


\begin{table}[t]
\centering
\small
\setlength{\tabcolsep}{5pt}
\begin{tabular}{@{} l c c @{}}
\toprule
\textbf{Method} & \textbf{Acc@T} & \makecell[c]{\textbf{Avg. Latency(s)}} \\
\midrule
Chain-of-Dictionary & 44.76 & 4.73 \\
Translate-and-Revise & 83.92 & 6.26 \\
DeepTrans & 41.24 & 3.10 \\
MT-R1-Zero & 40.43 & 3.79 \\
\textbf{Ours(BTG)} & \textbf{93.97} & \textbf{1.70} \\
\bottomrule
\end{tabular}
\vspace{-5pt}

\caption{Term accuracy and latency comparison against SOTA methods.}
\vspace{-20pt}

\label{tab:sota_comparison}
\end{table}

\subsection{Baseline Comparison}
To evaluate BTG's efficiency and precision, we compare it against prompt-based (Chain-of-Dictionary~\cite{lu-etal-2024-chain}), agent-based (Translate-and-Revise~\cite{pengcheng-etal-2024-translate}), R1-based (DeepTrans~\cite{wang-etal-2026-deeptrans}), and RL-based (MT-R1-Zero~\cite{feng-etal-2025-mt-r1}) paradigms. As Table~\ref{tab:sota_comparison} demonstrates, BTG achieves an optimal speed-precision balance. While agent-based models suffer from severe latency (6.26s) despite high accuracy (83.92), R1 and traditional RL approaches struggle with terminology alignment ($\text{Acc}@T < 42$), BTG establishes a new SOTA by securing peak semantic term accuracy (93.97) alongside the lowest average inference latency (1.70s).

\subsection{Data Quality Ablation}
\begin{figure}[t]
  \centering
  \includegraphics[width=1.0\columnwidth]{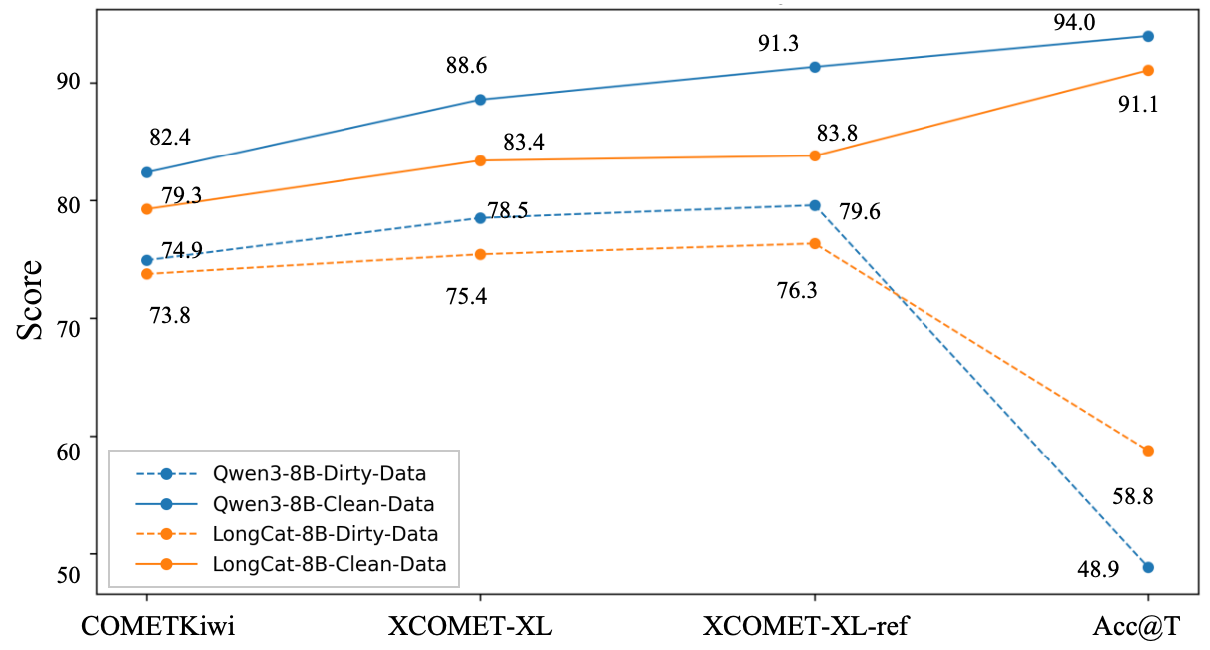}
    \vspace{-25pt}
  \caption{Ablation study on the effect of data quality.}
  \label{fig:data_quality}
    \vspace{-18pt}
\end{figure}
To demonstrate that our data cleaning method effectively enhances model training performance, we have conducted a data quality ablation study. As shown in Figure \ref{fig:data_quality}, we selected Qwen3-8B and LongCat-8B for comparative experiments. Across all metrics, models trained on cleaned data significantly outperformed those trained on the original data.
Most notably, Acc@T improved from 48.9 to 94.0 for Qwen3-8B, and from 58.8 to 91.1 for LongCat-8B. Other metrics also saw improvements of over 7 points relatively.  Furthermore, an interesting observation is that while Qwen3 showed weaker terminology capability than LongCat-8B on uncleaned data, it surpassed LongCat-8B when trained on the cleaned dataset.Overall, data quality is pivotal for enhancing LLMs' general and domain-specific translation.

\subsection{Data Mixing Ablation}
\label{sec:data_mixing_ablation}
To investigate the significant impact of data mixing ratios on model performance, we conduct two ablation experiments using reduced datasets.

\begin{figure}[t]
  \centering
  \includegraphics[width=1.0\columnwidth]{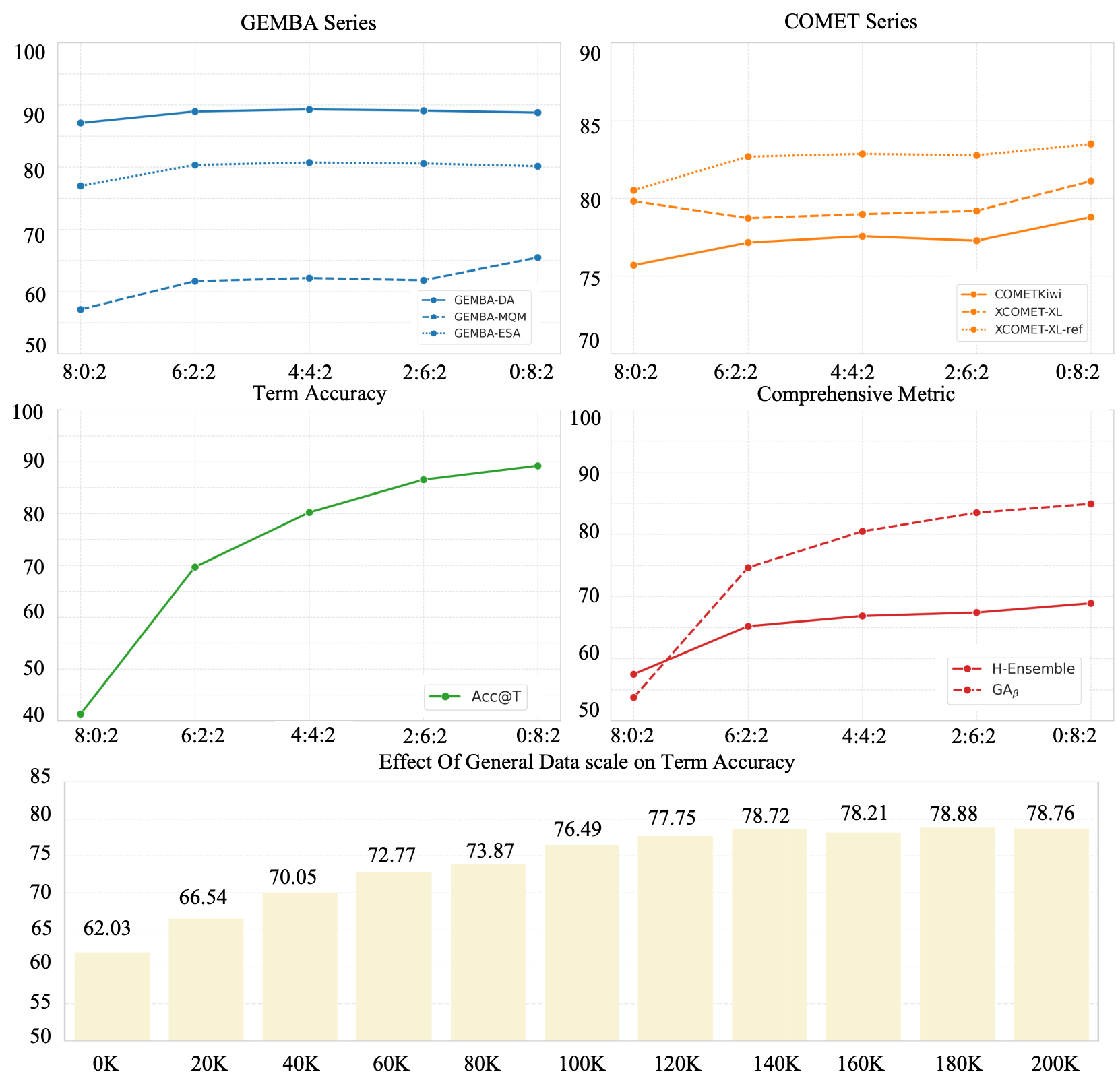}
    \vspace{-25pt}
\caption{Impact of data composition:(top) performance across General:Term:Instruction ratios, and (bottom) Acc@T scaling with general data under fixed term.}
  \label{fig:data_mixing}
    \vspace{-17pt}
\end{figure}

Under a fixed budget (Figure~\ref{fig:data_mixing}, top; 240k total, 20\% instruction data), shifting the general-to-terminology ratio from 8:0:2 to 0:8:2 significantly boosts Acc@T and composite metrics (H-Ensemble, GA$_{\beta}$). Stable GEMBA and COMET scores confirm enhanced domain specificity without sacrificing general fluency.

Fixing term data at 20k and scaling general samples to 200k (Figure~\ref{fig:data_mixing}, bottom) steadily improves Acc@T and plateaus at 140k. This indicates that general data synergistically boost term accuracy by reinforcing foundational grammar and semantics.

Combining these insights, we conclude that optimal performance demands both components: domain data ensures strict terminology adherence, while general data broadens overall semantic understanding. To pinpoint this ideal balance, we utilize the LLM Auto tool to derive the final data mixing ratio (Appendix~\ref{sec:llm_auto}).

\subsection{Instruction Following Experiment.}
\label{sec:instruction_following_exp}
We evaluate instruction-following on 2,000 samples across four dimensions (Figure~\ref{fig:instruction_paper}). Qwen3-8B-BTG successfully avoids the catastrophic forgetting typical of specialized models, maintaining comparable base-model performance in \textit{Style Transfer} (95.01 vs. 98.81) and \textit{Custom Terminology Injection} (55.23 vs. 56.21). Furthermore, BTG outperforms the baseline in adhering to rigid \textit{Structured Output} formats like JSON/HTML (94.43 vs. 92.62) and demonstrates superior adaptability with a significantly higher \textit{Correction Rate} for initial errors (74.32\% vs. 61.23\%). Thus, our framework combines vertical translation accuracy and general instruction-following (Figure~\ref{fig:instruction_paper}, Appendix~\ref{sec:instruction_appendix}).

\subsection{Online Human Evaluation}
\label{sec:Online_Human_Eval}
 
To assess deployment efficacy, experts blind-evaluated 1,568 business documents across four anonymized models: our SFT+GRPO framework (Model A), its W8A8 version (Model B), the in-house baseline (Model C), and an SFT-only ablation (Model D).

\begin{figure}[t]
  \centering
  \includegraphics[width=0.9\columnwidth]{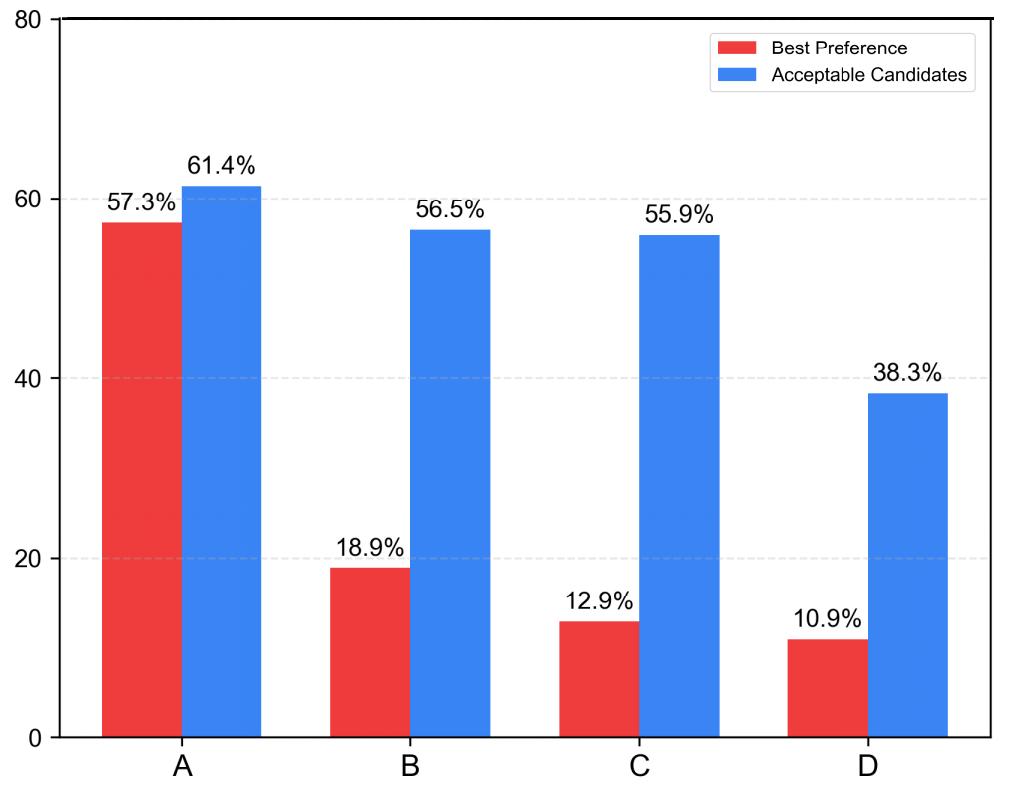}
    \vspace{-10pt}
  \caption{Expert evaluation of Best Preference and Acceptable Candidates on online documents.}
  \label{fig:online_human}
  \vspace{-20pt}
\end{figure}
As shown in Figure \ref{fig:online_human}, results show Model A overwhelmingly dominates in "Best Preference" at 57.3\% (vs. Model C's 12.9\% and Model D's 10.9\%), confirming that the GRPO stage is indispensable for human preference alignment. Furthermore, Model A leads the overall "Acceptance Rate" (61.4\%). Crucially, the quantized Model B (56.5\%) maintains an acceptance level comparable to the baseline (55.9\%), proving our framework is highly robust to quantization and exceptionally suited for resource-constrained online serving.

%% file: experiment/main_exp_table1.tex
\begin{table*}[t]
\centering
\small 
\renewcommand{\arraystretch}{0.95} 
\setlength{\tabcolsep}{2.5pt} 
\begin{tabular}{@{} l *{10}{c} @{}}
\toprule
\diagbox{Model}{Metric} & GM-DA & GM-MQM & GM-ESA & CM-Kiwi & XCM-XL & XCM-XL-ref & Acc@T & BLEU & H-Ensemble & GA$_{\beta}$ \\
\midrule
Qwen3-8B-BTG  & 95.55 & 76.75 & 90.95 & \underline{\textbf{82.38}} & \underline{\textbf{88.57}} & \underline{\textbf{91.35}} & \underline{\textbf{93.97}} & \underline{\textbf{41.78}} & \underline{\textbf{75.46}} & \underline{\textbf{92.43}} \\
Qwen3-8B      & 95.50 & 75.01 & 90.86 & 80.78 & 83.06 & 84.84 & 44.61 & 29.99 & 64.22 & 59.84 \\
Qwen2.5-7B    & 90.03 & 60.84 & 82.68 & 78.55 & 77.72 & 79.93 & 39.17 & 29.90 & 58.10 & 53.16 \\
Qwen3-Max     & 96.21 & 76.35 & 92.64 & 80.41 & 80.79 & 82.89 & 38.92 & 29.25 & 61.95 & 54.81 \\
Qwen-MT       & 93.89 & 76.67 & 90.99 & 81.86 & 84.59 & 85.56 & 47.33 & 31.07 & 65.55 & 62.27 \\
Deepseek-V3   & 96.39 & 77.70 & 92.94 & 81.00 & 82.52 & 84.77 & 49.13 & 31.31 & 66.21 & 64.28 \\
LongCat-8B    & 84.14 & 50.17 & 73.20 & 75.76 & 73.99 & 74.29 & 39.93 & 21.19 & 54.20 & 51.68 \\
LongCat-18B   & 85.37 & 53.28 & 75.52 & 77.07 & 75.57 & 76.14 & 38.48 & 22.27 & 55.02 & 50.99 \\
LongCat-Flash & 96.49 & 78.04 & 93.37 & 81.42 & 83.02 & 85.61 & 47.22 & 32.13 & 65.88 & 62.72 \\
GPT-4.1       & \underline{\textbf{96.64}} & \underline{\textbf{79.23}} & \underline{\textbf{94.02}} & 81.36 & 83.19 & 85.19 & 47.27 & 31.79 & 66.07 & 62.91 \\
\bottomrule
\end{tabular}
\vspace{-5pt}
\caption{Performance of BTG-tuned Qwen3-8B vs. 9 baselines across 10 metrics (GM: GEMBA, CM: COMET, XCM: XCOMET).}
\vspace{-15pt}
\label{tab:term_eval}
\end{table*}

%% file: appendix/training_dataset.tex
\begin{table}[H] 
    \centering
    \vspace{-7pt}
    \setlength{\tabcolsep}{3pt} 
    \resizebox{\columnwidth}{!}{%
    \begin{tabular}{@{} l r @{\hspace{4pt}} l r @{}}
        \toprule
        \multicolumn{2}{c}{\textbf{General Trans. Domain}} & \multicolumn{2}{c}{\textbf{Terminology Domain}} \\
        \cmidrule(r){1-2} \cmidrule(l){3-4}
        
        Zh-En-Pt Parallel Chat   & 1,676,758 & \begin{tabular}{@{}l@{}}Internet \& \\ Corp. Term.\end{tabular} & 229,718 \\
        
        United Nations Docs      & 1,084,881 & \multicolumn{2}{c}{\textbf{\textit{Instruction Following}}} \\ 
        \cmidrule(l){3-4} 
        
        Opus-100                 & 643,610   & Orca & 994,896 \\
        Literature and Film      & 329,077   & Belle & 519,255 \\
        Religion                 & 7,270     & Alpaca & 51,760 \\
        Corp. Encyclopedia       & 4,664     & & \\
        WMT24pp                  & 2,169     & & \\
        
        \midrule
        \multicolumn{3}{r}{\textbf{Grand Total}} & \textbf{5,544,058} \\
        \bottomrule
    \end{tabular}%
    } 
    \vspace{-8pt}
    \caption{Detailed statistics of training datasets.}
    \label{tab:data_stats}
    \vspace{-15pt}
\end{table}

%% file: appendix/llm_auto.tex
\begin{figure}[H]
  \centering
  \includegraphics[width=1.0\columnwidth]{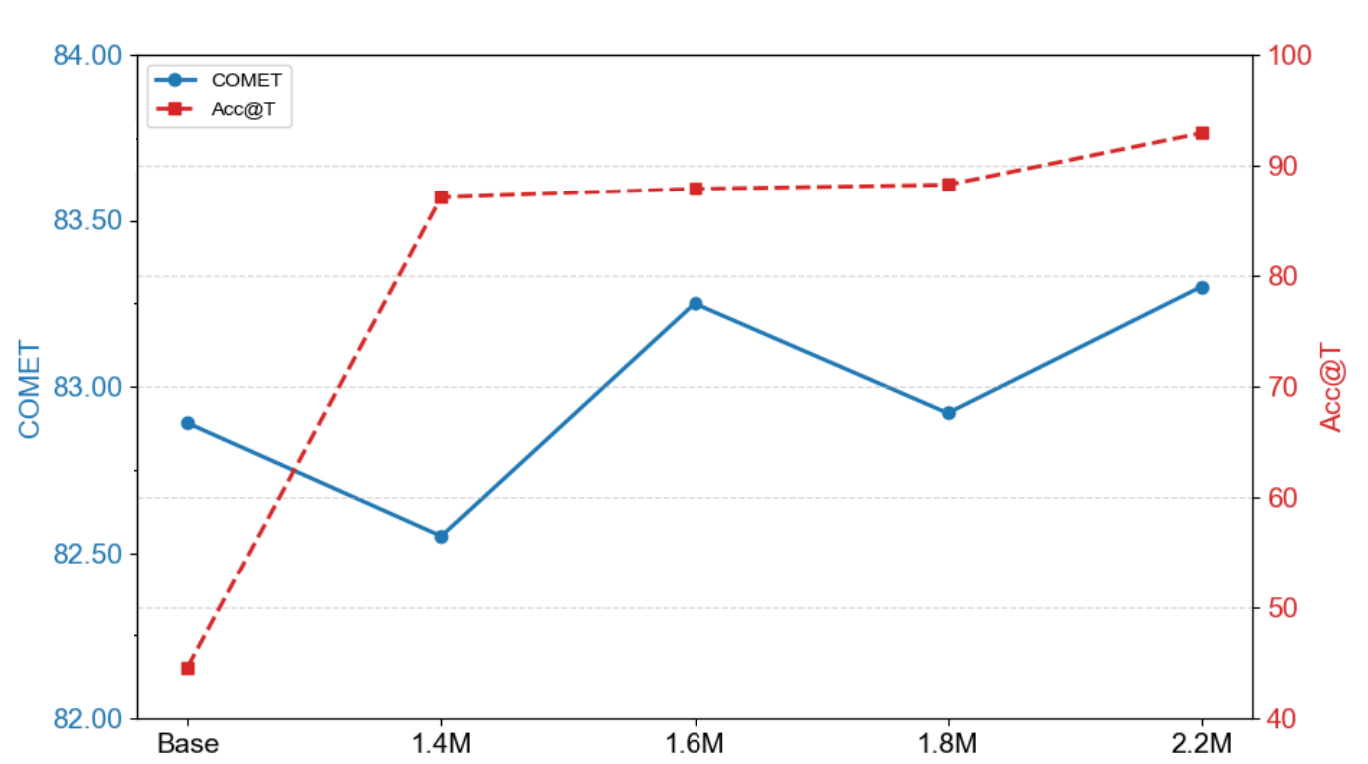}
    \vspace{-25pt}
  \caption{Performance trends under LLM-Driven data mixtures.}
  \label{fig:llm_search}
    \vspace{-20pt}
\end{figure}

We employ an iterative LLM-driven approach to optimize data composition based on feedback metrics. As shown in Figure \ref{fig:llm_search} and Tables \ref{tab:llm_suggest}-\ref{tab:llm_suggest4}, the model prioritized fixing terminology data at maximum capacity while progressively scaling up general data, eventually converging on the full corpus. Consistent with the findings in Section~\ref{sec:data_mixing_ablation}, general translation data enhances terminology accuracy to a certain extent.This trajectory confirms the feasibility and significant potential of LLM-driven strategies in devising effective data allocation schemes given domain insights.
\vspace{-10pt}

\begin{table}[H]
    \centering
    \begin{CJK*}{UTF8}{gbsn} 
    
    \begin{tcolorbox}[colback=gray!5, colframe=black!70, boxrule=1pt, arc=4pt, left=6pt, right=6pt, top=6pt, bottom=6pt]
        \small
        \textbf{System:} \\
\# Role \\
You are an LLM Data Mixing Optimization Expert. Your goal is to find the optimal training data ratio.\\
\# Objective\\
Find the best combination of General Data (N gen) and Terminology Data (N term) to:\\
1. Maximize Terminology Accuracy (Acc@T) > 44.61 (Base Model).\\
2. Keep General Capabilities (COMET) > 82.89 (Base Model).\\
\# Constraints\\
- Max N gen: 2,200,000\\
- Max N term: 229,718\\
- Step Size: Adjust N gen in units of 200000 or 400000.\\
\# Search Strategy (Prior Knowledge)\\
1. Anchor Law: High N term is required for high Acc@T. Prioritize values near the N term maximum.\\
2. Synergy Effect: Increasing N gen boosts Acc@T by improving linguistic ability until a saturation point.\\
\# Workflow\\
Analyze the provided history and suggest the next N gen and N term pair.\\
\# Output Format (JSON)\\
\{\\
    "analysis": "Brief analysis of trends",\\
    "recommendation": \{\\
        "general\_data\_count": <int>,\\
        "terminology\_data\_count": <int>\\
    \},\\
    "reasoning": "Why this specific combination?"\\
\}
    \end{tcolorbox}
    \end{CJK*}
    \caption{LLM-Driven mode system prompt}
    \label{tab:llm_suggest}
    \vspace{-20pt}
\end{table}

\begin{table}[H]
    \centering
    \begin{CJK*}{UTF8}{gbsn} 
    
    \begin{tcolorbox}[colback=gray!5, colframe=black!70, boxrule=1pt, arc=4pt, left=6pt, right=6pt, top=6pt, bottom=6pt]
        \small
        \vspace{0.5em}
        \textbf{User:} \\
Conduct a total of 4 rounds of experiments.\\
This is the first round of experiments.\\
History: [empty]\\
Please recommend the first starting detection point.\\
        
        \vspace{0.5em}
        \textbf{Assistant:} \\
\{\\    "analysis": "With no prior history, we must select a starting point that balances both objectives. Since maximizing terminology accuracy requires high N term, starting near its maximum is logical. For N gen, beginning with a mid-to-high value allows us to gauge the synergy effect without risking immediate saturation or underperformance in general capabilities.",\\"recommendation": \{        "general\_data\_count": 1400000,        "terminology\_data\_count": 229718    \}, \\"reasoning": "This combination leverages the Anchor Law by maximizing N term to ensure strong terminology accuracy. Choosing 1,400,000 for N gen is a strategic midpoint: it\'s substantial enough to support general capabilities (likely above the COMET threshold) and provides room to explore both upward and downward adjustments in future rounds. This setup gives a clear signal for both objectives and sets a strong baseline for further optimization."\}\\
    \end{tcolorbox}
    \end{CJK*}
    \caption{LLM-Driven mode 1st round}
    \label{tab:llm_suggest2}
    \vspace{-20pt}
\end{table}

\begin{table}[H]
    \centering
    \begin{CJK*}{UTF8}{gbsn} 
    
    \begin{tcolorbox}[colback=gray!5, colframe=black!70, boxrule=1pt, arc=4pt, left=6pt, right=6pt, top=6pt, bottom=6pt]
        \small
        \vspace{0.5em}
        \textbf{User:} \\
Round: 1\\
Num info: ["general\_data\_count": 1400000, "terminology\_data\_count": 229718]\\
Evaluation result:\\
- COMET: 82.55\\
- Acc@T: 87.09\\
        
        \vspace{0.5em}
        \textbf{Assistant:} \\
\{\\    "analysis": "The first experiment achieved high terminology accuracy (Acc@T: 87.09) by maximizing N\_term, but COMET (82.55) is slightly below the base model threshold (82.89). This suggests that while N\_term is optimal, N\_gen needs to be increased to improve general capabilities without sacrificing terminology accuracy. The synergy effect indicates that boosting N\_gen should enhance COMET and may further support Acc@T up to a saturation point.",\\    "recommendation": \{        "general\_data\_count": 1600000,        "terminology\_data\_count": 229718    \},\\    "reasoning": "Increasing N\_gen to 1,600,000 (next step up) while keeping N\_term at its maximum leverages the synergy effect to improve COMET, aiming to surpass the threshold. Since Acc@T is already high, maintaining N\_term ensures continued terminology performance. This adjustment is incremental, allowing us to monitor the impact on both metrics and avoid overshooting the optimal balance."\}\\

    \end{tcolorbox}
    \end{CJK*}
    \caption{LLM-Driven mode 2rd round}
    \label{tab:llm_suggest23}
    \vspace{-20pt}
\end{table}

\begin{table}[H]
    \centering
    \begin{CJK*}{UTF8}{gbsn} 
    
    \begin{tcolorbox}[colback=gray!5, colframe=black!70, boxrule=1pt, arc=4pt, left=6pt, right=6pt, top=6pt, bottom=6pt]
        \small
        \vspace{0.5em}
        \textbf{User:} \\
Round: 2\\
Num info: ["general\_data\_count": 1600000, "terminology\_data\_count": 229718]\\
Evaluation result:\\
- COMET: 83.25  - Acc@T: 87.88\\
        \vspace{0.5em}
        \textbf{Assistant:} \\
\{\\    "analysis": "Increasing N gen to 1,600,000 improved both metrics: COMET rose above the threshold (83.25 > 82.89), and Acc@T increased to 87.88. This confirms the synergy effect, where more general data boosts both general and terminology performance. However, the gains in Acc@T are starting to plateau, indicating we may be approaching saturation for terminology accuracy. Further increases in N gen may continue to benefit COMET, but the improvement in Acc@T may be marginal.",\\    "recommendation": \{        "general\_data\_count": 1800000,        "terminology\_data\_count": 229718    \},\\    "reasoning": "To fully explore the synergy effect and ensure COMET remains robust, increasing N gen to 1,800,000 is a logical next step. Keeping N term at its maximum maintains terminology accuracy. This will help determine if further increases in general data yield meaningful improvements or if saturation is reached, providing valuable information for fine-tuning the optimal ratio."\}
        
    \end{tcolorbox}
    \end{CJK*}
    \caption{LLM-Driven mode 3rd round}
    \label{tab:llm_suggest3}
    \vspace{-20pt}
\end{table}
\vspace{-50pt}

\begin{table}[H]
    \centering
    \begin{CJK*}{UTF8}{gbsn} 
    
    \begin{tcolorbox}[colback=gray!5, colframe=black!70, boxrule=1pt, arc=4pt, left=6pt, right=6pt, top=6pt, bottom=6pt]
        \small
        \vspace{0.5em}
        \textbf{User:} \\
Round: 3\\
Num info: ["general\_data\_count": 1800000, "terminology\_data\_count": 229718]\\
Evaluation result:\\
- COMET: 82.92 - Acc@T: 88.25\\
        \vspace{0.5em}
        \textbf{Assistant:} \\
\{\\    "analysis": "With N gen at 1,800,000, COMET remains above the threshold (82.92), and Acc@T continues to improve (88.25), though the gains are smaller, indicating diminishing returns for terminology accuracy. The synergy effect is still present, but Acc@T is nearing saturation. COMET is stable and consistently above the base model. This suggests we are close to the optimal balance, but there may be room for a final test at the upper limit of N gen.",\\    "recommendation": \{        "general\_data\_count": 2200000,        "terminology\_data\_count": 229718    \},\\    "reasoning": "Testing the maximum allowed N gen (2,200,000) with maximum N term will confirm whether Acc@T and COMET plateau or if there is any further improvement. This final experiment will help establish if the highest general data volume offers the best synergy or if previous rounds already represent the optimal mix. It ensures we have fully explored the parameter space within constraints."\}
        \vspace{0.4em}
        \hrule 
        \textbf{Final Result:} \\
Round: 4\\
Num info: ["general\_data\_count": 2200000, "terminology\_data\_count": 229718]\\
Evaluation result:\\
- COMET: 83.98\\
- Acc@T: 92.94
        
    \end{tcolorbox}
    \end{CJK*}
    \caption{LLM-Driven mode 4th round and final result}
    \label{tab:llm_suggest4}
    \vspace{-20pt}
\end{table}

%% file: appendix/term_extract_prompt.tex
\subsubsection{Terminology Extraction Prompt}

Table~\ref{tab:extraction_prompt} presents the semantic-aware terminology mining prompt. Here, \texttt{[DOMAIN]} denotes the user-defined domain, \texttt{[LANG1/2]} are the detected languages, and \texttt{[SENT1/2]} represent the parallel sentence pair from the corpus.

As illustrated in Table \ref{tab:example_extract}, we present a complete example of extracting terminology pairs from a dialogue. Specifically, within the food delivery domain, the Chinese term \begin{CJK*}{UTF8}{gbsn}"步兵"\end{CJK*} is appropriately translated as "the on foot" in English.
\vspace{-25pt}
\begin{table}[H]
\begin{table}[H]
\centering
\begin{tcolorbox}[colback=gray!10, colframe=black!70, boxrule=1pt, arc=4pt, left=6pt, right=6pt, top=6pt, bottom=6pt]
\small
\textbf{System Instruction:} \\
You are a professional multilingual terminology extractor.

\vspace{0.2cm}
\textbf{Input:} \\
Given two semantically equivalent sentences in different languages, 
extract aligned key term pairs.
Only extract terms related to \texttt{<DOMAIN>}.
Sentence 1 (\texttt{<LANG1>}): \texttt{<SENTENCE1>} \\
Sentence 2 (\texttt{<LANG2>}): \texttt{<SENTENCE2>}

\vspace{0.2cm}
\textbf{Output Requirements:} \\
Output format: \texttt{\{"pairs": [\{"<LANG1>": "...", "<LANG2>": "..."\}]\}} \\
Only output valid JSON. If no clear term alignment, output \texttt{\{"pairs": []\}}.
\end{tcolorbox}
\caption{Prompt template for terminology extraction.}
\label{tab:extraction_prompt}
\vspace{-10pt}
\end{table}

\centering
\begin{tcolorbox}[colback=gray!10, colframe=black!70, boxrule=1pt, arc=4pt, left=6pt, right=6pt, top=6pt, bottom=6pt]
\begin{CJK*}{UTF8}{gbsn} 

\small
\textbf{System Instruction:} \\
You are a professional multilingual terminology extractor.

\vspace{0.2cm}
\textbf{Input:} \\
Given two semantically equivalent sentences in different languages, 
extract aligned key term pairs.
Only extract terms related to \texttt{" Food Delivery "}.

\vspace{0.1cm}

\begin{tabularx}{\linewidth}{@{} l X @{}}
    Sentence 1 (Chinese): & {\ttfamily 该站点急招CBD商圈的步兵，主要负责午高峰期间写字楼内部的送餐服务，无需自备电动车。} \\ 
    \addlinespace[0.5em] 
    Sentence 2 (\texttt{English}): & {\ttfamily We are seeking the on foot for the CBD to handle intra-building deliveries during the midday peak.}
\end{tabularx}

\vspace{0.2cm}
\textbf{Output Requirements:} \\
Output format: \texttt{\{"pairs": [\{"Chinese": "...", "English": "..."\}]\}} \\
Only output valid JSON. If no clear term alignment, output \texttt{\{"pairs": []\}}.

\vspace{0.2cm}
\textbf{Model Output:} \\
\texttt{\{"pairs": [\{"Chinese": "步兵", "English": "the on foot"\}]\}}
\end{CJK*}

\end{tcolorbox}
\caption{An example of extracting practical terms, extracting Chinese ``\begin{CJK*}{UTF8}{gbsn}步兵\end{CJK*}'' and English "the on foot" from similar sentences}
\label{tab:example_extract} 
\vspace{-20pt}
\end{table}

\subsubsection{Domain adaptability of term mining}

To demonstrate the generalizability and robustness of our semantic-aware terminology mining framework across diverse fields, we extend our empirical evaluation beyond the cross-border corporate domain to two highly specialized disciplines: Biomedicine and Agriculture. Specifically, for the medical domain, we constructed the corpus by crawling 16 sets of topic-aligned multilingual news and articles from the official websites of the World Health Organization (WHO) and the United Nations (UN). Similarly, for the agricultural domain, we gathered 5 sets of relevant multilingual documents from the Food and Agriculture Organization (FAO).

Table~\ref{tab:domain_generalization} outlines the comprehensive statistics of these additional domains, detailing the volume of initially mined and subsequently augmented term pairs, the final scale of the training sets, and the downstream performance gains ($\text{Acc}@T$). The significant accuracy improvements observed across both the medical domain ($41.84 \rightarrow 70.41$) and the agricultural domain ($43.77 \rightarrow 64.33$) robustly confirm that our proposed methodology is highly adaptable and can be seamlessly transferred to disparate specialized domains without requiring architectural modifications.

\begin{table}[htbp]
\centering
\small
\setlength{\tabcolsep}{4pt} 
\begin{tabular}{@{} l c c c c @{}} 
\toprule
\multicolumn{1}{c}{\textbf{Domain}} & \textbf{Mined} & \textbf{Aug.} & \textbf{Train} & \textbf{Acc@T} \\
 & \textbf{Pairs} & \textbf{Pairs} & \textbf{Set} & \\
\midrule
Biomedicine & 828 & 8,073 & 40K & 41.84 $\rightarrow$ 70.41 \\
Agriculture & 295 & 2,841 & 20K & 43.77 $\rightarrow$ 64.33 \\
\bottomrule
\end{tabular}
\caption{Statistics and performance across specialized domains. The significant gains in $\text{Acc}@T$ demonstrate cross-domain generalizability.}
\label{tab:domain_generalization}
\end{table}

%% file: appendix/cross_lingual_Evaluation.tex
\begin{table}[htbp]
    \centering
    \small
    \setlength{\tabcolsep}{3pt} 
    \resizebox{\columnwidth}{!}{%
    \begin{tabular}{@{} l cccc @{}}
        \toprule
        \textbf{Model} & \textbf{CMKiwi} & \textbf{XCM-XL} & \textbf{XCM-ref} & \textbf{Acc@T} \\ 
        \midrule
        \multicolumn{5}{c}{\textit{English $\rightarrow$ Portuguese}} \\
        \midrule
        Qwen3-8B-BTG & \underline{\textbf{84.56}} & \underline{\textbf{92.82}} & \underline{\textbf{94.43}} & \underline{\textbf{95.36}} \\
        Qwen3-8B & 83.27 & 86.69 & 86.74 & 50.63 \\
        Qwen2.5-7B & 79.72 & 80.40 & 80.54 & 41.65 \\
        Qwen3-Max & 81.99 & 83.68 & 84.69 & 50.98 \\
        Qwen-MT & 84.24 & 88.49 & 88.99 & 54.34 \\
        Deepseek-V3 & 83.02 & 87.11 & 87.91 & 56.75 \\
        LongCat-8B & 76.55 & 75.89 & 75.85 & 44.71 \\
        LongCat-18B & 78.70 & 78.09 & 78.01 & 41.05 \\
        LongCat-Flash & 83.92 & 88.03 & 88.50 & 54.47 \\
        GPT-4.1 & 83.14 & 87.56 & 88.40 & 55.46 \\
        \midrule
        \multicolumn{5}{c}{\textit{Portuguese $\rightarrow$ Chinese}} \\
        \midrule
        Qwen3-8B-BTG & \underline{\textbf{81.13}} & \underline{\textbf{79.84}} & \underline{\textbf{86.09}} & \underline{\textbf{93.48}}\\
        Qwen3-8B & 80.34 & 71.40 & 79.04 & 34.43 \\
        Qwen2.5-7B & 79.17 & 69.83 & 77.06 & 33.16 \\
        Qwen3-Max & 77.38 & 63.38 & 74.19 & 34.79 \\
        Qwen-MT & 81.06 & 73.15 & 81.89 & 38.45 \\
        Deepseek-V3 & 80.17 & 71.72 & 81.24 & 40.95 \\
        LongCat-8B & 78.19 & 68.05 & 76.07 & 35.09 \\
        LongCat-18B & 79.45 & 70.71 & 78.37 & 35.35 \\
        LongCat-Flash & 80.84 & 72.22 & 81.20 & 39.04 \\
        GPT-4.1 & 81.06 & 73.02 & 81.58 & 37.04 \\
        \bottomrule
    \end{tabular}%
    }
    \vspace{-6pt}
    \caption{Cross-lingual breakdown analysis of the main experiment. Top: English to Portuguese. Bottom: Portuguese to Chinese. (CM: COMET)}
    \label{tab:benchmark_results_crosslang}
    \vspace{-15pt}
\end{table}

To comprehensively assess the model's multilingual capabilities in frequently used cross-border languages, we provide a detailed performance breakdown for specific translation directions. 

As presented in Table \ref{tab:benchmark_results_crosslang}, the lower side
 displays the scores of each model for Portuguese to Chinese translation across various metrics. Notably, our model achieves scores significantly higher than other baselines on xCOMET-XL, xCOMET-XL-ref, and Acc@T, indicating superior performance in both general translation capability and terminology translation for the Portuguese-to-Chinese pair. 

The upper side of the table shows the comparison for English to Portuguese translation. Similarly, our model secures high scores exceeding 90 on these three key metrics, with Acc@T reaching 95.36. This highlights our outstanding performance in English-to-Portuguese translation. These experimental results demonstrating that our proposed method significantly enhances the model's understanding and translation capabilities across frequently used cross-border cooperation languages.

%% file: appendix/inference_efficiency.tex
\begin{table}[t]
    \centering
    \small 
    \setlength{\tabcolsep}{3pt} 
    \resizebox{\columnwidth}{!}{
    \begin{tabular}{l cc cc cc}
        \toprule
        Length & \multicolumn{2}{c}{TTFT (s)} & \multicolumn{2}{c}{Gen Time (s)} & \multicolumn{2}{c}{Hallu. (\%)} \\
        \cmidrule(lr){2-3} \cmidrule(lr){4-5} \cmidrule(lr){6-7}
        (Tokens) & BTG & Base & BTG & Base & BTG & Base \\
        \midrule
        0--1.5k    & 0.41 & 11.66 & 12.26 & 12.32 & 3.96 & 9.63 \\ 
        1.5k--3k   & 0.50 & 13.78 & 21.74 & 21.22 & 6.34 & 10.47 \\  
        3k--5k     & 0.63 & 16.39 & 30.43 & 29.24 & 5.96 & 12.08 \\ 
        \bottomrule
    \end{tabular}%
    }
    \vspace{-5pt}
    \caption{Inference efficiency comparison across different input lengths (Hallu:Hallucination Rate ).}
    \label{tab:latency}
    \vspace{-15pt}
\end{table}

To evaluate efficiency advantages of our approach over traditional prompt-based injection, we conducted a latency stress test using real-world office documents ranging from 0 to 5,000 tokens. We benchmarked Qwen3-8B-BTG against the Qwen3-8B base model, which relies on lengthy prompts to inject terminology context. As presented in Table \ref{tab:latency}, our model achieves significant TTFT reduction (e.g., 0.63s vs. 16.39s for long docs), successfully avoiding the prompt processing overhead that causes the base model's latency to surge. Additionally, mean generation times remain comparable, confirming that our domain adaptation introduces no additional inference overhead. Finally, our model significantly lowers hallucination rates (e.g., 5.96\% vs. 12.08\% in the 3k-5k range); unlike long-context prompting which risks attention distraction, our approach ensures greater stability and instruction adherence.

%% file: appendix/training_method_albation_study.tex
\begin{figure}[t]
  \centering
  
  \includegraphics[width=1.0\columnwidth]{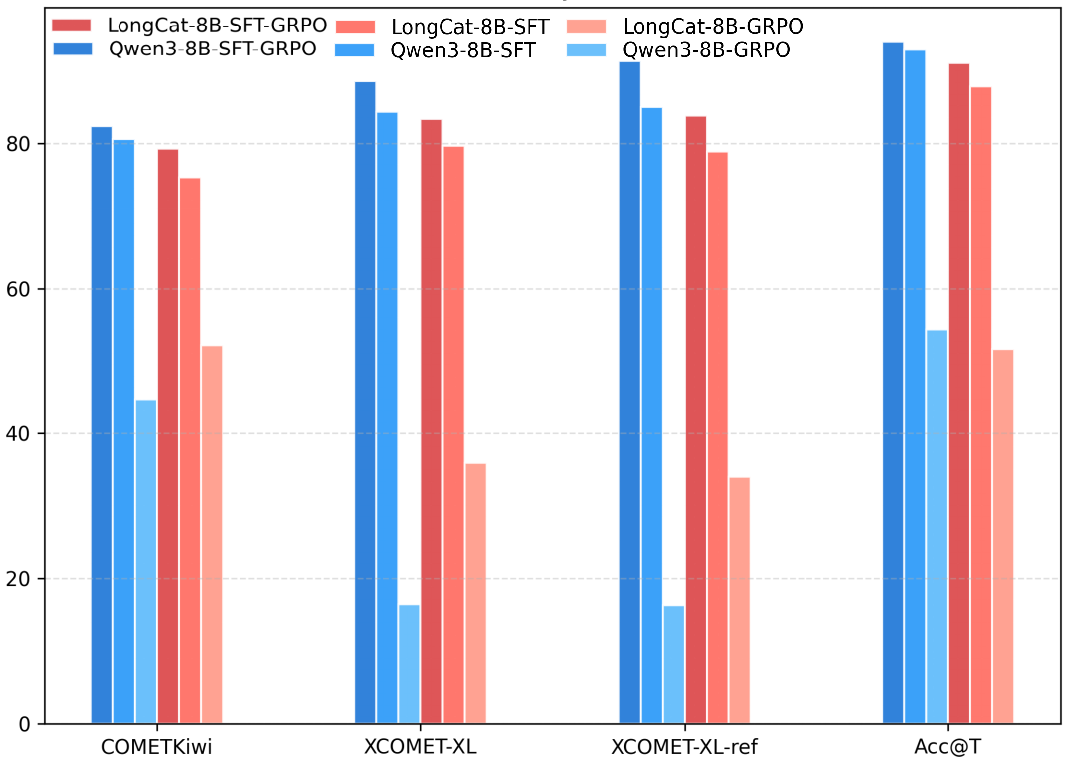}
    \vspace{-20pt}
  \caption{Impact of training methods on model performance.}
  \label{fig:train_method}
    \vspace{-20pt}
\end{figure}
To justify the necessity of our proposed two-stage training paradigm, we conducted an ablation study on Qwen3-8B and LongCat-8B by isolating the training stages. We compared three variants: (1) GRPO-Only, (2) SFT-Only, and (3) SFT+GRPO (our full two-stage method).

As illustrated in Figure\ref{fig:train_method}, both models exhibit a consistent trend across all metrics. 

\textbf{Key Findings 1: Failure of RL-only.} GRPO-Only models perform poorly, scoring significantly lower across all dimensions. This indicates that reinforcement learning alone is insufficient for injecting domain terminology. Lacking a supervised foundation, the model struggles to explore the correct terminology space, leading to suboptimal convergence.

\textbf{Key Findings 2: SFT Establishes a Strong Baseline.} The SFT-Only models show a substantial performance leap, proving that supervision data is the primary driver for knowledge acquisition. However, they still lag behind the full method.

\textbf{Key Findings 3: Synergy of SFT and GRPO.} The best performance is consistently achieved by SFT+GRPO two-stage models. For instance, on the critical Acc@T metric, Qwen3-8B-SFT-GRPO surpasses its SFT-only counterpart. This confirms our hypothesis: SFT effectively "teaches" model vertical knowledge, while the subsequent GRPO stage acts as a refinement mechanism, aligning model's output preferences to be more natural and contextually appropriate, thereby unlocking full potential of injected knowledge.


%% file: appendix/instruct_follow.tex
\begin{figure*}[t]
  \centering
  \includegraphics[width=2.0\columnwidth]{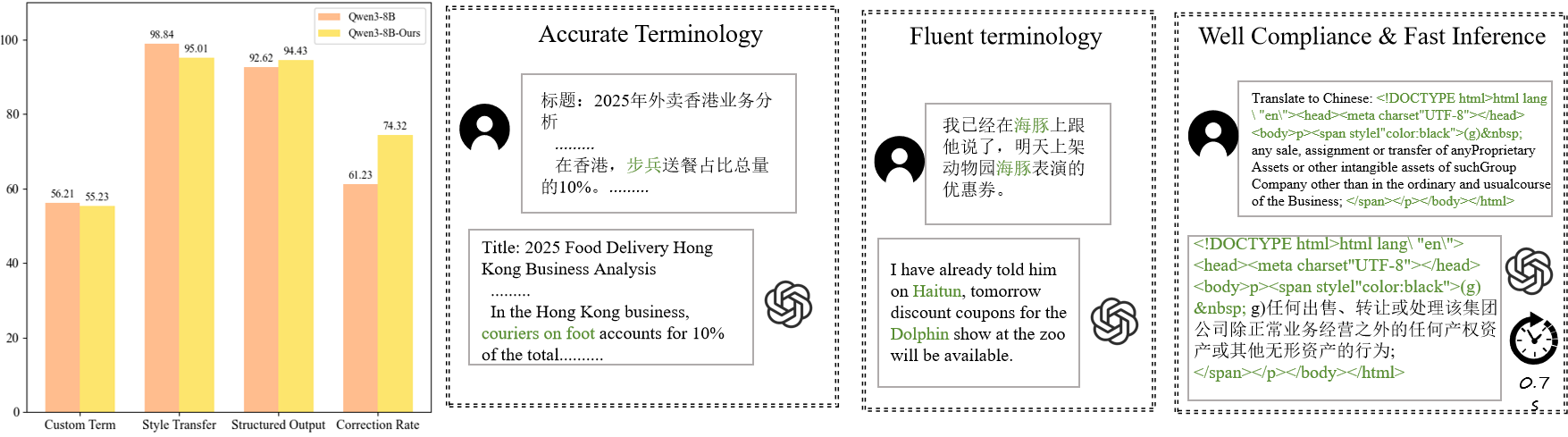}
  \caption{Performance of trained vs. base models on 4 tasks: correction, structured output, style transfer, and custom terminology.}
    \vspace{-10pt}
  \label{fig:instruction_paper}
\end{figure*}
We evaluated the model's instruction-following capabilities on 2,000 samples from online user documents across four dimensions: \textit{Style Transfer}, \textit{Custom Terminology Injection}, \textit{Structured Output} (JSON/HTML), and \textit{Correction Rate} (success rate in correcting initial errors).

In Figure \ref{fig:instruction_paper}, our trained model (Qwen3-8B-BTG) maintains robust general capabilities compared to base model.
For Style Transfer and Custom Terms, our model achieves scores of 95.01 and 55.23 respectively, remaining comparable to the base model (98.81 and 56.21). This confirms that our methods did not cause catastrophic forgetting.
In Structured Output, our model (94.43) outperforms the base model (92.62), demonstrating superior adherence to rigid syntax constraints.
Notably, our model achieves a 74.32\% correction rate when handling feedback on initial errors, significantly surpassing the base model (61.23\%), indicating higher adaptability.The right-hand three pictures in Figure~\ref{fig:instruction_paper} show corresponding examples; for more examples, please refer to \ref{sec:instruction_appendix}

Specialized translation models often face degradations in their general instruction-following capabilities; in contrast, our framework successfully combines vertical precision with the versatile instruction-following nature of general LLMs.

The following four real-world cases demonstrate the model's performance in multi-turn dialogues subject to specific requirements, including error correction, structured output generation, style transfer, and adherence to user-defined terminology.

Table~\ref{tab:custom_injection} illustrates a dynamic scenario where the model incorporates user-specified translations. It strictly adheres to these constraints while preserving syntactic fluency and semantic coherence, demonstrating robust instruction-following capabilities.

As shown in Table~\ref{tab:correction_example}, the model initially translated \begin{CJK*}{UTF8}{gbsn}"下沉市场"\end{CJK*} as ``underdeveloped market'' instead of the correct term ``the lower-tier market.'' Upon receiving the user's correction, the model successfully revised the translation. Notably, all correction examples are drawn from failure cases encountered in the first round of our main experiments.

In Table \ref{tab:structured_output}, our model is able to accurately translate text containing HTML structures while fully preserving the integrity of HTML format. This demonstrates that our model can effectively adhere to structured output requirements.

As shown in Table \ref{tab:style_tansfer}, this is an experiment on style transfer in which the user requests a more detailed translation. After understanding the user's requirements, the model revises its output to produce a more comprehensive and detailed translation.
\vspace{-10pt}
\begin{table}[H]
    \centering
    \begin{CJK*}{UTF8}{gbsn} 
    
    \begin{tcolorbox}[colback=gray!5, colframe=black!70, boxrule=1pt, arc=4pt, left=6pt, right=6pt, top=6pt, bottom=6pt]
        \small
        \textbf{User:} \\
        Translate to English: \\
        把今天的会议总结一下，发到他的邮箱里面，并记得抄送给我

        \vspace{0.5em}
        \textbf{Assistant:} \\
        Summarize today's meeting and send it to his email, and remember to copy me.

        \vspace{0.8em}
        \hrule 
        \vspace{0.8em}
        \textbf{User:} \\
        把抄送翻译为CC\\
        \textit{(Trans: Translate 抄送 to cc.)}
        
        \vspace{0.5em}
        \textbf{Assistant:} \\
        Summarize today’s meeting, send it to his email, and remember to CC me.

    \end{tcolorbox}
    \end{CJK*}
    \caption{A task example of user custom term injection}
    \label{tab:custom_injection}
\end{table}
\begin{table}[H]
    \centering
    \begin{CJK*}{UTF8}{gbsn} 
    
    \begin{tcolorbox}[colback=gray!5, colframe=black!70, boxrule=1pt, arc=4pt, left=6pt, right=6pt, top=6pt, bottom=6pt]
        \small
        \textbf{User:} \\
        Translate to English: \\
        下沉市场过去常被主流平台忽略，但随着消费力崛起和数字化渗透，正成为拉动增长的新引擎。
        
        \vspace{0.5em}
        \textbf{Assistant:} \\
        The \textcolor{red}{\textit{underdeveloped market}} was often overlooked by mainstream platforms in the past, but with the rise of consumer purchasing power and digital penetration, it is now emerging as a new engine driving growth.
        
        \vspace{0.8em}
        \hrule 
        \vspace{0.8em}
        
        \textbf{User:} \\
        下沉市场应该翻译为 ``The lower-tier market''.\\
        \textit{(Trans: Please translate 下沉市场 as ``The lower-tier market''.)}
        
        \vspace{0.5em}
        \textbf{Assistant:} \\
        The \textcolor{blue}{\textbf{lower-tier market}} was often overlooked by mainstream platforms in the past, but with the rise of consumer purchasing power and digital penetration, it is now emerging as a new engine driving growth.
        
    \end{tcolorbox}
    \end{CJK*}
    \caption{An example of the terminology correction task.}
    \label{tab:correction_example}
    \vspace{-20pt}
\end{table}

\begin{table}[H]
    \centering
    \begin{CJK*}{UTF8}{gbsn} 
    
    \lstset{
        basicstyle=\ttfamily\small,
        breaklines=true,       
        columns=flexible,
        frame=none,            
        extendedchars=false,   
        escapechar=`           
    }
    
    \begin{tcolorbox}[colback=gray!5, colframe=black!70, boxrule=1pt, arc=4pt, left=6pt, right=6pt, top=6pt, bottom=6pt]
        
        \textbf{User:} \\
        Translate to English: 
        \begin{lstlisting}
        <!DOCTYPE html>
        <html lang\"zh\"><head>
        <meta charset\"UTF-8\"></head><body>
        <p><span style\"color: black\">(g) \&nbsp;
最新推出的大模型支持“边写边译”功能，用户在输入中文时可实时生成高质量多语种译文，提升跨境沟通效率。 </span></p>
        </body> </html>
        \end{lstlisting}
        \textbf{Assistant:} 
        \begin{lstlisting}
        <!DOCTYPE html>
        <html lang"en"><head>
        <meta charset"UTF-8"></head><body>
        <p><span style"color: black">(g) \&nbsp;
The newly launched large model supports the 'Translate as You Type' feature, enabling users to generate high-quality multilingual translations in real-time while inputting Chinese, thereby enhancing cross-border communication efficiency.； </span></p>
        </body></html>
        \end{lstlisting}
    \end{tcolorbox}
    \end{CJK*}
    \caption{An example of structured output translation task.}
    \label{tab:structured_output}
\end{table}

\begin{table}[H]
    \centering
    \begin{CJK*}{UTF8}{gbsn} 
    
    \begin{tcolorbox}[colback=gray!5, colframe=black!70, boxrule=1pt, arc=4pt, left=6pt, right=6pt, top=6pt, bottom=6pt]
        \small
        \textbf{User:} \\
        Translate to English: \\
        在外卖行业,步兵并不是指军队里面的步兵，而是一支深入城市街巷、高效执行落地任务的一线团队，包括骑手，他们是连接平台与用户的关键力量。
        
        \vspace{0.5em}
        \textbf{Assistant:} \\
        In the food delivery industry, couriers on foot do not refer to infantry in the military, but rather to an efficient frontline team that penetrates urban alleys and streets to execute delivery tasks, including riders who are the key force connecting the platform with users.
        \vspace{0.8em}
        \hrule 
        \vspace{0.8em}
        \textbf{User:} \\
        请进行详细解释翻译，以便对方理解。\\
\textit{(Trans: Please explain the translation in detail so that the other party can understand.)}
        
        \vspace{0.5em}
        \textbf{Assistant:} \\
        In the food delivery industry, couriers on foot are not the same as soldiers on foot. Instead, they form a highly efficient and effective frontline team that delivers orders directly to customers in urban alleys and streets. This team of couriers plays a crucial role in connecting food delivery platform with its users.
    \end{tcolorbox}
    \end{CJK*}
    \caption{An example of sytle transfer translation task.}
    \label{tab:style_tansfer}
\end{table}